# Keeping it simple: Implementation and performance of the proto-principle of adaptation and learning in the language sciences

Petar Milin
*University of Birmingham, UK*

Harish Tayyar Madabushi
*University of Birmingham, UK*

Michael Croucher
*Numerical Algorithms Group, UK*

Dagmar Divjak
*University of Birmingham, UK*

**Abstract**: In this paper we present the Widrow-Hoff rule and its applications to language data. After contextualizing the rule historically and placing it in the chain of neurally inspired artificial learning models, we explain its rationale and implementational considerations. Using a number of case studies we illustrate how the Widrow-Hoff rule offers unexpected opportunities for the computational simulation of a range of language phenomena that make it possible to approach old problems from a novel perspective.

**Funder acknowledgment:** Petar Milin and Dagmar Divjak were supported by a Leverhulme Trust Research Leadership Award (RL-2016-001).

## 1. Introduction

One of the earliest neurally inspired models of learning is the Widrow-Hoff (1960) rule. It is predated by three publications only: the seminal work of McCulloch and Pitts (1943) that hypothesized how neurons might work by relying on analogy to electrical circuits; Donald Hebb's book *The Organization of Behavior* (1949), which famously stipulated the basic principle of association of neurons by means of neural co-activation (i.e., assembling); and Frank Rosenblatt's work on the *Perceptron* (Rosenblatt, 1958). Importantly, however, the Widrow-Hoff rule was the first one that was successfully applied to real-life problems (e.g., noise cancellation in telephone lines which is used to date; cf., Haykin, 1999).

After the initial excitement and until the (more) recent successes, models such as those mentioned above that were inspired biologically or, more specifically, neurally were ignored in favour of machines implementing von Neumann's traditional architecture. During the 1970s, the Widrow-Hoff rule was accidentally re-discovered in Psychology by Rescorla and Wagner (1972) who worked on animal and human learning, and by Kohonen (1972) in his work on Self-Organizing Maps in Computer Science. Finally, the widely known and successful Connectionist Parallel-Distributed Processing Models have the Widrow-Hoff rule as their principal building block (cf., McClelland & Rumelhart, 1986).

The Widrow-Hoff learning rule is often termed the Delta rule or the Least Mean Square (LMS) rule. And as the last and most descriptive name – Least Mean Square – indicates, the rule aims at minimizing (the mean square of) the difference between the targeted and the actual data or the response. This embodies the Principle of Minimal Disturbance by "[reducing] the output error for the current training pattern, with minimal disturbance to responses already learned" (Widrow & Lehr, 1990, p. 1423), that is where "in the light of new input data, the parameters of an adaptive system should only be disturbed in a minimal fashion" (Haykin, 1996, p. 436).

In what follows we will present the rule in more detail. This is followed by a discussion on the implementational challenges posed by the need to model learning over large datasets, with many inputs and outputs. The final part presents some case studies which will serve to illustrate the main points about the implementation and performance of this basic adaptive learning principle in the realm of the language sciences.

## 2. ADAptive LINear Element(s) – ADALINE and MADALINE

As mentioned above, the Widrow-Hoff rule is the 'soft 'part of many other neurally inspired learning systems (i.e., machines). Initially, it was developed to govern the 'hard 'part of a system called ADALINE – ADAptive LINear Element. In Widrow and Hoff's own words, ADALINE is essentially "an adaptive pattern classification machine […] for the purpose of illustrating adaptive behavior and artificial learning" (Widrow & Hoff, 1960, p. 97). MADALINE, similarly, stands for Multiple ADAptive LINear Elements (Widrow, 1962; Widrow & Lehr, 1990), which aligne more than one ADALINE to give a combined output by using various threshold logic functions (e.g., AND, OR, MAJORITY). Over 20 years, (M)ADELINEs have had a range of successful applications (e.g., high-speed modem control: Lucky, 1965; telephone/satellite noise canceler: Sondhi, 1967; signal processing/filtering: Widrow & Stearns, 1985; Widrow, Mantey, Griffiths, & Goode, 1967; adaptive control: Widrow, 1987).

(M)ADALINE shows adaptive behaviour by predicting the next state of the environment-element (or system of elements) dyad and, as the rule prescribes, this is achieved by minimizing the average number of erroneous predictions, by means of the Least Mean Square error (i.e., the difference). The process is iterative to allow for fine adaptive changes towards ever-better performance over time. In other words, examples of input and desired output are fed to the system in a step-by-step (discrete) manner. As the experience accrues, the system's competence accrues too – the more examples, the better the performance (Widrow, 1959; Widrow & Hoff, 1960; Widrow & Lehr, 1990; Haykin, 1996, 1999). Generally speaking, however, this is possible only if there is *systematicity* in experience; i.e., the input data and the targeted outcome are related and, moreover, that relationship can be discerned or inferred statistically/probabilistically (cf., Widrow, 1959; Widrow & Lehr, 1990; for more recent discussions see Chen, Haykin, Eggermont, & Becker, 2008; Milin, Nenadić, & Ramscar, under review).

Widrow drew several important conclusions about ADALINE's performance from his own statistical theory of adaptation (Widrow, 1959, 1960). The most important one is that the quality of adaptation depends only on the number of examples seen; i.e., the amount of experience. However, experimental work in animal and human learning and, in particular, the formal model of learning of Rescorla and Wagner (1972), pointed out some additional important properties of this particular type of *error-correction learning* (Haykin, 1999). First, the underline{background rate} of informative (associated or correlated) and non-informative examples crucially affects the learning performance (cf., Rescorla, 1968). Second, the order of examples can also alter the learning inasmuch that initially learned relations can underline{block} those that follow (Kamin, 1969). (For more details about these properties of error-correcting learning see Rescorla & Wagner, 1972; Rescorla, 1988, 2003; Arnon & Ramscar, 2012; Ramscar, Yarlett, Dye, Denny, & Thorpe, 2010; Milin, Divjak, & Baayen, 2017).

With this in mind, we can conclude that the 'simple' ADALINE is, in fact, surprisingly robust and rather complex and sophisticated. Being neurally inspired, it comes as no surprise that (M)ADALINEs and the Widrow-Hoff rule (or its psychologically motivated 'incarnation' in the rule of Rescorla and Wagner), are considered to be among a handful of biologically plausible learning principles (see Chen et al., 2008; and similar points made by Rescorla & Wagner, 1972; Rescorla, 1988; Enquist & Ghirlanda, 2005), and even that this particular rule might have evolutionary advantages over more complex learning principles (cf., Trimmer, McNamara, Houston, & Marshall, 2012).

Even though Widrow-Hoff represents a foundational principle of many other adaptive learning systems, it was harshly criticized together with other similar rules (like Rosenblatt's perceptron) by Minsky and Papert (1988). They claimed that these early models have limited theoretical value as their applicability is limited to linearly separable domains or problem spaces. We would, however, agree with the counterargument of Widrow and Lehr (1990) that "Adaline's inability to realize all functions is in a sense a strength rather than the fatal flaw envisioned by some critics of neural networks (Minsky & Papert, 1988), because it helps limit the capacity of the device and thereby improves its ability to generalize" (p. 1422). The point here is that one must consider a trade-off between a network's capacities (a) to master the classification challenges and (b) to be able to generalize to patterns that were not present during learning. Widrow and Lehr, themselves, conclude (in a somewhat sardonic tone) that "if generalization is not needed, we can simply store the associations in a look-up table, and will have little need for a neural network" (p. 1422).

Interestingly enough, logistic regression, which was introduced as early as 1840 but became popular in its current form only in the late 60s (Cramer, 2002) through the work of Cox 1966 (Cox, 1966), deals with problems that are not linearly separable. Similarly, Artificial Neural Networks, introduced in 60s and later popularised by Rumelhart, Hinton and Williams (cf., Goodfellow, Bengio, & Courville, 2016), also deal with data that is not linearly separable. As predicted by Widrow and Lehr, however, both of these methods suffer from *overfitting*, a phenomenon whereby the model fits too closely to the training data and hence fails to generalise. While there are methods of getting around the issue of overfitting, such as regularisation, it is notable that Widrow and Lehr's (1990) argument was, thus, conclusively validated.

So, what is this rule of Widrow and Hoff that is 'limited 'but capable to generalize, and seems to capture a core property of learning and adaptation?

## 3. The rule of learning

As stated above, the Widrow-Hoff rule aims to minimize the mean square difference between the predicted (expected) and the actual (observed) data or response. In the authors 'own words "the design objective is the minimization of the average number of errors" (Widrow & Hoff, 1960, p. 96). Yet, thinking in terms of the average number of errors might lead to the false conclusion that the method evaluates its results 'after the fact', i.e., when all the data is available for estimating the average number of errors. Instead, as is observed in learning in biological organisms from which this method ultimately draws inspiration, the process is gradual: data becomes available in a piece-meal fashion and learning by minimizing the number of errors also 'evolves 'over time, in an iterative fashion. How, then, does the error become minimal?

To answer this question, we will make use of a typical neurally inspired processing unit. Its schematic presentation is as follows:

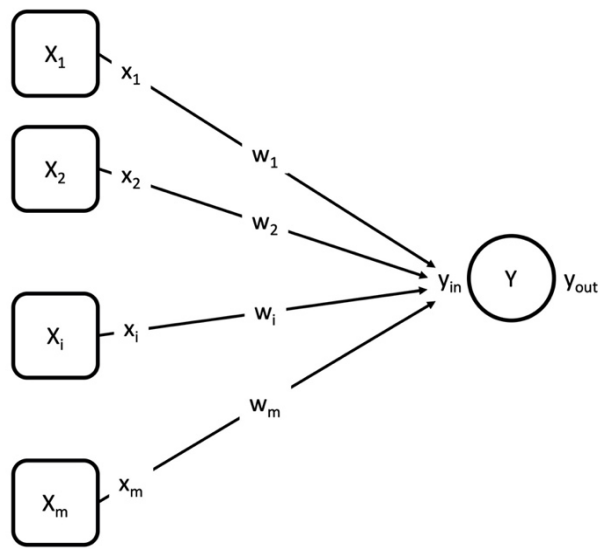

where $X_i$ is the *i*-th input unit, with the $x_i$ signal that is weighted (or scaled) by the weight $w_i$ and transmitted to the output unit $Y$, whose net (or total) input is $y_{in}$ while $y_{out}$ is its activation (output signal). Essentially, this schema represents a biological neuron with its main elements: dendrites (input units, as $X_i$), soma (as $Y$), and axon (with output signal, as $y_{out}$).

As the Principle of Minimal Disturbance prescribes, the reduction in output error takes place at the time when new input data arrives to the $X_i$ units and is sent as weighted signal, $x_i w_i$, to be summed up into the net input signal, $y_{in}$, of $Y$:

$$y_{in} = x_1 w_1 + x_2 w_2 + \cdots + x_n w_n = \sum_{i=1}^{n} x_i w_i \qquad (1)$$

Crucially, for each new input signal, $x_i$, the weights, $w_i$, are changed (or "disturbed" in Haykin's 1996 words, p. 436) minimally, given the difference between the *criterion* (the target or the teacher) of learning, $T$, and the current net input, $y_{in}$: $(T - y_{in})$. This difference is also weighted by the *learning rate* parameter, $\gamma$, a conveniently small value that guarantees that learning is gradual (for some discussion see, for example, Widrow & Hoff, 1960; Rescorla & Wagner, 1972; Blough, 1975; Enquist, Lind, & Ghirlanda, 2016):

$$\Delta w_i = x_i \gamma (T - y_{in}) \qquad (2)$$

Again, the presence or absence of a new input signal, $x_i$, will make change happen or not, while the learning rate, $\gamma$, will ensure *minimal disturbance* to what is already learned by comparing the desired state of the signal, $T$, with the current net signal, $y_{in}$. The change in weight, $\Delta w_i$, will update the weight, $w_i$, itself, as the states in time before ($t$) and after ($t + 1$) the new input arrived:

$$w_i^{t+1} = w_i^t + \Delta w_i^t \qquad (3)$$

More technically, the Least Mean Square (LMS) is applied at the current time, and given the input signal, which make it a *stochastic gradient descent* method. It is a gradient descent as it *iteratively* seeks for the *local minimum* of an objective function of LMS. Also, it is stochastic since it is estimated on random samples of the data. The assumption is that the error is Gaussian distributed, uncorrelated and with zero mean. Widrow and Hoff (1960) argued that rule performance is highly resilient to violations of this distributional assumption and is affected only if the correlation exceeds 0.8.

In conclusion of this section, it is worth noting that the Widrow-Hoff rule of learning has two theoretically important generalizations: the Kalman Filter (Kalman, 1960) and the Temporal Difference model (Sutton & Barto, 1990). Without going into much detail, for the current purposes only, we will briefly comment on the specifics of these generalizations (for details, however, we refer to the original works, i.e., to Sutton, 1992; while for pioneering work in integrating these independent lines of research and for more recent overviews to, e.g., Kruschke, 2008; Gershman, 2015).

Strictly speaking, LMS is equivalent to the maximum likelihood estimator of the weight. Thus, it is a point-estimate that does not take into account any uncertainty in the signal; i.e., the *expectation* of the weight is given but not its *variability* (although, as we discussed, the error is assumed to be Gaussian). Furthermore, for the Widrow-Hoff rule, the learning rate is predetermined and static (constant). Somewhat simplified, the Kalman (1960) Filter generalizes the basic Widrow-Hoff rule by explicitly taking into account the uncertainty present in the signal and, by updating that uncertainty too, it dynamically and item-specifically adjusts the rate of learning, at each learning event. As the learning rule takes both the expectation and the variance into account, the learning rate parameter becomes the Kalman Gain, which directly and positively correlates with the uncertainty that drives the process of iterative learning.

Finally, Widrow-Hoff learning is driven exclusively by the difference between the criterion and the current net input. Future, longer-term 'encounters 'or, in the terminology of Reinforcement Learning (RL: cf., Niv, 2009; Gureckis & Love, 2015), future reward expectations are not taken into account. Temporal Difference learning (Sutton & Barto, 1990) takes into account future predictions rather than only immediate ones. That update rule is similar to the Widrow-Hoff rule but extends it by adding the future (reward) expectation term to the criterion (the target, $T$) of learning. However, if that term is set to zero, the update equation becomes identical to the updates of the Widrow-Hoff rule.

## 4. Implementation of the Widrow-Hoff rule

A set of input cues and corresponding outcomes can be encoded as two matrices, $\mathbf{C}$ and $\mathbf{T}$, which directly correspond to Widrow-Hoff input and output units ($X_j$ and $Y_k$). Each row of these matrices, jointly, represents a learning event. Hence, matrix $\mathbf{C}$ has $i$ rows (number of events) and $j$ columns (number of cues) and matrix $\mathbf{T}$ has the same number of rows, $i$, and $k$ columns (number of outcomes).

To train the model we begin by initialising the weights matrix, $\mathbf{W}$, to zero. This matrix has $j$ rows and $k$ columns (the given number of cues and outcomes, respectively). Every learning event, $i$, is defined by the two row vectors of the current cues ($\mathbf{c} = \mathbf{C}_{[i,]}$) and the current outcomes ($\mathbf{t} = \mathbf{T}_{[i,]}$). To update the weight matrix, we apply the Widrow-Hoff equations (2) and (3), in matrix form:

$$\mathbf{W}_{\text{new}} = \mathbf{W}_{\text{old}} + \mathbf{c} \cdot \gamma \cdot \left( \mathbf{t} - \mathbf{W}_{\text{old}}^{T} \cdot \mathbf{c} \right) \tag{4}$$

where $\mathbf{W}_{\text{old}}$ is the weight matrix before update, $\mathbf{W}_{\text{old}}^{T}$ is the transpose of weight matrix before update, and $\gamma$ is the learning rate. $\mathbf{W}_{\text{new}}$ is the weight matrix after the update.

As is the case with Deep Learning algorithms, the majority of the computation time is spent performing linear algebra operations such as vector/matrix multiplication. Such operations are highly parallelisable and modern computers have several features that allow them to perform these calculations extremely quickly.

To make optimal use of modern hardware it is necessary to ensure that the algorithm makes use of matrix and vector operations (so-called 'vectorisation of code') to the extent possible and that the computational environment (e.g., **R**, **Python**, **Julia** etc.) uses a strong implementation of BLAS (Basic Linear Algebra Subroutines). In this section we will briefly describe the speeds achieved on a standard laptop and an HPC (High Performance Computing) server using both **R** and **MATLAB** implementations. The former is a massively popular environment for statistical computing while the latter leads in highly optimized BLAS implementations.

Multiple CPU cores provide coarse-grained parallelisation so one would expect a typical 4 core laptop to be able to speed up an operation by up to a factor of 4. In addition to coarse-grained parallelisation, current CPUs offer options for fine-grained parallelism through SIMD (Single Input Multiple Data) vector instructions and FMA (Fused-Multiply-Add) routines that allow each CPU core to perform operations such as element-wise multiplication or addition of vectors on 8 or more floating points per clock cycle. Together these two levels of parallelism (e.g., 4 cores × 8 SIMD × 2 FMA) deliver many operations per clock cycle. Combined with additional hardware-level tricks such as Cache Tiling, which maximises the use of the small cache of high-performance memory that sits inside a CPU, it is possible to achieve huge speed-ups compared to naive implementations.

Many libraries make standard use of these opportunities. For example, the matrix multiplication ($\mathbf{C = A\% * \%B}$) in **R** relies on the **Fortran** or **C** libraries using BLAS. The efficiency of the BLAS library depends on the details of the implementation (e.g., OpenBLAS, ATLAS or IntelMKL). For example, on one of our Windows laptops, multiplying two 2000 × 2000 matrices together in **R** takes 5.5 seconds using the CRAN version of **R** that uses a reference BLAS, and 0.25 seconds using **Microsoft R Open** which uses the highly optimised IntelMKL implementation of the BLAS. This is a speed difference of over 20 times with no difference in **R** code.

Yet, even with a fast BLAS, **R** is not the most efficient language for matrix-based computations as it does not offer the option of using GP-GPUs (General Purpose-Graphics Processing Units) and single precision arithmetic. **MATLAB** (MATrix LABoratory) uses a strong BLAS (IntelMKL) as default and offers many other optimisations for vector and matrix arithmetic. On the same Windows laptop, our **MATLAB** implementation of the Widrow-Hoff rule is up to 2 (1.866) times faster than our **R** implementation. In addition, **MATLAB** allows the use of GP-GPU which were originally developed for graphics processing. However, their architecture makes them highly suited for numerical linear algebra and they are the computational force that has made modern Deep Learning computationally viable.

A third avenue for speed-gain that we are currently exploring relies on single precision arithmetic. Single precision numbers use 32bits compared to 64bits for double precision and so they use less memory and are faster to compute with. This comes at the price of reduced precision, however, and care needs to be taken to ensure that the answers obtained using single precision are correct.[1]

Finally, further gains in speed have been identified in cases when the matrices of cues and outcomes are very sparse – i.e., when they are coded as present or absent (1 or 0). Our current sparse implementation on a CPU is just over 2.5 times slower than our GPU version. However, since CPUs have access to much more memory and the sparse implementation requires less memory, the sparse implementation is probably the most useful for large-scale problems. A full sparse matrix algorithm is not possible because the weights matrix becomes dense after even just one iteration of Widrow-Hoff rule.

To sum up, in our performance investigations, we have used a random sample of 15,000 utterances from the Touchstone Applied Science Associates (TASA) corpus (Ivens & Koslin, 1991). Each utterance presented one learning event. As is typical for many of the Naive

---

[1] Note that we have not yet used single precision in any of the research simulations presented in this paper and have also not investigated when the single precision versions of our algorithm might become unstable. We have, however, noted that there is significant speed up for both CPUs and GPUs and so further investigation would be useful.

Discrimination Learner (NDL, presented in greater detail in Section 5.1, below) simulations that have been successfully used in addressing a range of language-specific research problems, the input cues were letter triplets and the outcomes were word forms occurring in sampled utterances. Across 15,000 learning events (i.e., utterances) there were 5799 cues and 8775 outcomes. We trained a Widrow-Hoff learning network in a single simulation run (i.e., no repeated runs on the same data or learning events), using both **R** and **MATLAB** implementations on various platforms. **MATLAB** was used for single precision and GPU runs since our **R** implementation does not support these. The results are summarized in Table 1.

 Our tests thus show that we achieved an almost 120 times faster performance compared to the reference **R** implementation, while keeping calculations in double precision. To test the performance on a large-scale dataset, we made use of the whole TASA corpus from which the rarest character strings (possible typographic errors, oddities and such) had bene removed. This test dataset consisted of 42,695 cues and 42,695 outcomes (both word forms), distributed over 10,761,965 learning events. In other words, both input cue and outcome matrices ($\mathbf{C}$ and $\mathbf{T}$) contained $10,761,965 \times 42,695$ rows and columns, while the weight matrix ($\mathbf{W}$) had $42,695 \times 42,695$ rows and columns. Such a large-scale problem represents a challenge for a GPU-based approach as the resulting weight matrix would be approximately 13GB, which is too large for the standard GPU unit to handle. Our sparse implementation in **MATLAB**, however, completed in approximately 30 hours on a 32 core HPC server (2x Intel Xeon E5-2698 v4).[2]

**Table 1.** Time (in seconds) taken for a single run of Widrow-Hoff rule on the TASA sample.

| Software/Hardware | Single precision | Double precision |
|---|---|---|
| CRAN R 3.6.2 on 4 core laptop (Intel i7-8650U) | N/A | 5725.6 |
| Microsoft R Open 3.6.2 on 4 core laptop (Intel i7-8650U) | N/A | 4260.1 |
| MATLAB 2019a on 4 core laptop (Intel i7-8650U) | 1554.2 | 3068.5 |
| MATLAB 2019a on 32 core HPC server (2x Intel Xeon E5-2698 v4) | 290.5 | 540.4 |
| MATLAB 2019a with laptop GPU (Geforce GTX 1050) | 186.6 | 370.9 |
| MATLAB 2019a sparse implementation on 32 core HPC server (2x Intel Xeon E5-2698 v4) | N/A | 130.7 |
| MATLAB 2019a with server GPU (Tesla V100) | 26.2 | 48.8 |

---

[2] We have also identified further opportunities for optimising the sparse version which will be released at a later date.

**5. Applications in the language domain**

Computationally modelling simulated data is gaining in importance across many disciplines, including psychology and linguistics: it provides in-depth understanding of a particular aspect or behaviour of a complex system at a relatively low cost. Furthermore, the results of computational modelling can inspire the specific experimental manipulations, as well as make it possible to derive informed and specific hypotheses that can be tested empirically. The Widrow-Hoff rule seems to be a particularly well-suited for these tasks due to its simplicity and its biological plausibility (cf., Chen et al., 2008; Trimmer et al., 2012; Enquist et al., 2016). In addition to that, being a building block of more complex models (e.g., Kohonen's 1972 Self-Organizing Maps, Connectionist PDP Models as proposed by McClelland & Rumelhart, 1986), the Widrow-Hoff learning principle allows us to draw more far-reaching inferences about the power that these complex model might bring. Yet, one needs to keep in mind that, as models become increasingly more powerful and 'capable', their complexity is paid for by *intractability* and questionable *evolutionary prerogatives* (cf., Trimmer et al., 2012). Ultimately, then, the question would be: how far can we go and how well can we do if we assume that learning proceeds in a simple error-correction fashion, as the Widrow-Hoff rule assumes?

5.1. Previous applications to language phenomena

Although the lay man would think it is utter madness to claim that language is not learned (what else could it be?), language scientists have long shied away from considering learning-based approaches to language cognition. Yet, recently a number of papers have appeared that redress the balance and take an explicit learning-based approach to language processing. This was made possible through the implementation of the Rescorla-Wagner rule as the Naive Discrimination Learner (NDL) by Baayen, Milin, Đurđević, Hendrix, and Marelli (2011). NDL provides a computational framework for error-driven discrimination learning and is, essentially, identical to the algorithm presented in this paper (cf. Rescorla, 2008).

      The network structure of NDL models is very simple: the nodes in the layer of cues are linked up to nodes in the layer of outcomes and both cues and outcomes are symbolic representations. There are no intervening hidden layers. In a seminal paper, Baayen et al. (2011) successfully accounted for paradigmatic effects on the processing of case inflections in Serbian using only letter unigrams and bigrams (one- or two-letter sequences) as cues, and representations of lexical, inflectional or derivational meanings as outcomes. Moreover, they simulated the effects of frequency, family size, and contextual effects on the processing of simple words, inflected words, derived words, pseudo- derived words, compounds and prepositional phrases in English. In other words, using a simple two-layer network the authors demonstrated how known morphemic, lexical and phrasal effects observed in naming, decision and reading tasks arise without the need for morphemic, lexical and phrasal representations.

Milin, Feldman, Ramscar, Hendrix, and Baayen (2017) expanded the basic orthography-based NDL architecture and work with two discrimination networks: a grapheme-to-lexome (G2L) network that links trigraphs (cues consisting of three-letter sequences) to lexomes (outcomes), and a lexome-to-lexome (L2L) network in which lexomes are both cues and outcomes. In both cases, lexomes are pointers to locations in high-dimensional semantic vector space. Several measures can be derived from the G2L and L2L networks. The G2L matrix yields information about the extent to which the target lexome is activated by orthography, the amount of uncertainty with which an outcome is associated given a specific cue, and the availability of a lexome irrespective of input. The L2L matrix yields indications of a lexome's semantic density, its semantic typicality, its prior availability and the extent to which other lexomes are co-activated. These discrimination-based measures of lexical processing outperformed classical lexical-distributional measures, in particular, frequency counts and measures of form similarity (e.g., neighbourhood size and density), in accounting for the behavioural data as collected in priming studies.

Lexomes capture experiences, including linguistic experiences, that are discriminated by a speech community. The combination of an orthographic and a semantic network consisting of linguistic and non-linguistic category labels (such as 'motion 'or 'past') successfully captured (self-paced) reading behaviour in Russian and explained how different types of learners use orthography and semantics to guide their reading (Milin, Divjak, et al., 2017).

So far, the naïve discrimination-learning measures have been used successfully as predictors of various behaviours, as measured through naming, decision and reading tasks across a range of European, Semitic and Asian languages. NDL is attractive to language scientists because it is an algorithm that is psychologically and neurobiologically plausible and yields patterns that are learnable from experience. Learning is driven by positive and negative experiences, and in this spirit, NDL captures both evidence in favour of and against associations between cues and outcomes. As such, it goes beyond what can be achieved with frequency counts.

In what follows we will briefly present three case studies to explore further how the Widrow-Hoff rule can be used for a variety of problems in language research. We will start with a simple simulation study questioning how humans acquire colour labels (Section 5.2). Next, we will test whether Widrow-Hoff learning weights can help predict specific linguistic abstractions given behavioural responses (Section 5.3). Finally, we will use Widrow-Hoff weights as embeddings for the deep learning machine (Section 5.4).

## 5.2. Learning to name basic colours

It is usual in computational simulations to make certain simplifying assumptions. For our illustration of how Widrow-Hoff learns to 'name 'basic colours, we started from the normalized spectral sensitivity of cone cells in the human eye (as described by Dowling, 1987; and later used by Schubert, 2006), presented in Figure 1. Next, as the Normal distribution is defined by its mean ($M$; position parameter) and standard deviation ($sd$; scale parameter), with 99.73% of values between −3 and +3 $sd$ from the mean, we defined 12 probability regions using steps of half a standard deviation, from $-\infty$ to $+\infty$, as illustrated in Figure 2.

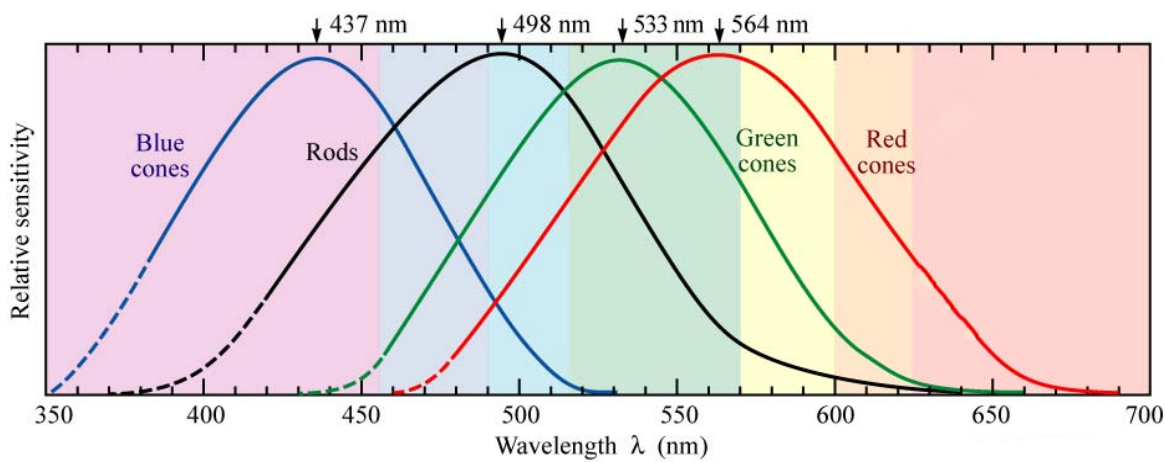

**Figure 1.** Normalized spectral sensitivity of cone cells in the human eye (adopted from Dowling, 1987 and Schubert, 2006).

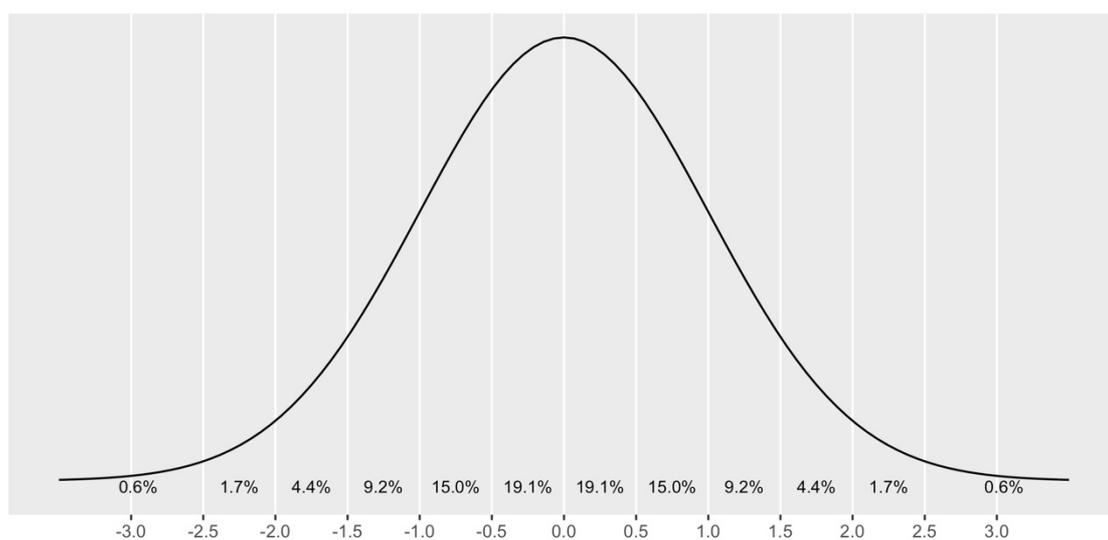

**Figure 2.** Standard Normal curve with half a standard deviation units and respective percentages under the curve.

The normalized spectral sensitivity of cone cells and the probability under the Normal distribution, combined, allow us to define 12 ranges of spectral sensitivity (in nanometres – $nm$), for each of the three types of cone cells (blue, green, and red), as presented in Table 2.

**Table 2.** Limits of normalized relative sensitivity for three types of cone cells per half a standard deviation unit.

| | BLUE | GREEN | RED |
|---|---|---|---|

| | *sd* | lo. limit | up. limit | lo. limit | up. limit | lo. limit | up. limit |
|---|---|---|---|---|---|---|---|
| 1 | (-∞, -2.5] | 350.0 | 364.5 | 428.0 | 445.5 | 450.0 | 469.0 |
| 2 | (-2.5, -2.0] | 364.5 | 379.0 | 445.5 | 463.0 | 469.0 | 499.0 |
| 3 | (-2.0, -1.5] | 379.0 | 393.5 | 463.0 | 480.5 | 499.0 | 507.0 |
| 4 | (-1.5, -1.0] | 393.5 | 408.0 | 480.5 | 498.0 | 507.0 | 526.0 |
| 5 | (-1.0, -0.5] | 408.0 | 422.5 | 498.0 | 515.5 | 526.0 | 545.0 |
| 6 | (-0.5, 0.0] | 422.5 | 437.0 | 515.5 | 533.0 | 545.0 | 564.0 |
| 7 | (0.0, 0.5] | 437.0 | 451.5 | 533.0 | 550.5 | 564.0 | 583.0 |
| 8 | (0.5, 1.0] | 451.5 | 466.0 | 550.5 | 568.0 | 583.0 | 602.0 |
| 9 | (1.0, 1.5] | 466.0 | 480.5 | 568.0 | 585.5 | 602.0 | 621.0 |
| 10 | (1.5, 2.0] | 480.5 | 495.0 | 585.5 | 603.0 | 621.0 | 640.0 |
| 11 | (2.0, 2.5] | 495.0 | 509.5 | 603.0 | 620.5 | 640.0 | 659.0 |
| 12 | (2.5, +∞] | 509.5 | 524.0 | 620.5 | 638.0 | 659.0 | 678.0 |

The training data for the Widrow-Hoff learning rule consisted of 100,000 randomly generated wavelength datapoints, in the range of the visible spectrum (i.e., from 350 $nm$ to 750 $nm$, which means that we added 50 $nm$ below and above the typically assumed visible range). The individual draw was equally probable in that given range of wavelengths. Then, each generated value was tested for falling into a particular cone cell sensitivity range, and the corresponding probability under the Normal curve was assigned as numeric learning input cue for each type of cone cells, i.e., the respective relative sensitivity per cone cell type (see Figure 1).

For example, the first random draw gave a value of 493.74 $nm$. This means that the value falls in 10th value range for the blue cells, in the 4th for the green cells, and in the 2nd for the red cells. The respective relative sensitivities are, thus, 0.044, 0.092, and 0.017 (for blue, green, and red cones). The input cue-outcome combination for that learning event would, then, consist of the three probability values for the three cone cells and an indicator of the correct colour label, given what a human participant would typically report to see (consult Figure 1 and its coloured zones). For our previous example, the learning event would be:

| | CUES | | | OUTCOMES | | |
|---|---|---|---|---|---|---|
| | **blue** | **green** | **red** | **blue** | **green** | **red** |
| | 0.044 | 0.092 | 0.017 | 0 | 0 | 0 |

Note that, in this particular case, all outcomes are false (0) as the value of 493.74 $nm$ points to a wavelength that is typically reported as turquoise, not as blue or green. However, had the randomly drawn value been 462.26 $nm$, it would be typically reported as blue, and then the learning event would be as follows:

| CUES | | | OUTCOMES | | |
|------|------|------|------|------|------|
| blue | green | red | blue | green | red |
| 0.150 | 0.017 | 0.006 | 1 | 0 | 0 |

We have trained the Widrow-Hoff update rule on the generated dataset with 100,000 learning events, while setting the learning rate parameter to $\gamma = 0.1$. The updating of associative weights is summarized in Figure 3. The figure shows very different developmental profiles for each of the three colour words and their respective weights for three cone-specific relative sensitivities, as learned by the Widrow-Hoff update rule. Of course, this is a simplification of the actual complexity of the perceptual process, as we take into account physical input and the colour label only. Nevertheless, the results are instructive.

Overall, these different profiles of associations between cues (probabilities) and outcomes (labels) clearly allow for each label to be learned. The word "red" (Figure 3, right panel) is triggered when the activation for the red cones is positive, for the blue cones is neutral, and for the green cones is negative. For the word "green" (Figure 3, middle panel), conversely, both red cones and green cones are positively activated, although the green cones are more strongly activated and there is a moderate 'suppression' of red cone activation over time (i.e., it shows a slow decrease in weight values). At the same time, the blue cones have a mild negative association with the word "green". Finally, the Widrow-Hoff rule learns that, to name the colour "blue" (Figure 3, left panel), blue and green cones are almost equally positively activated, with a strong negative association with red cones.

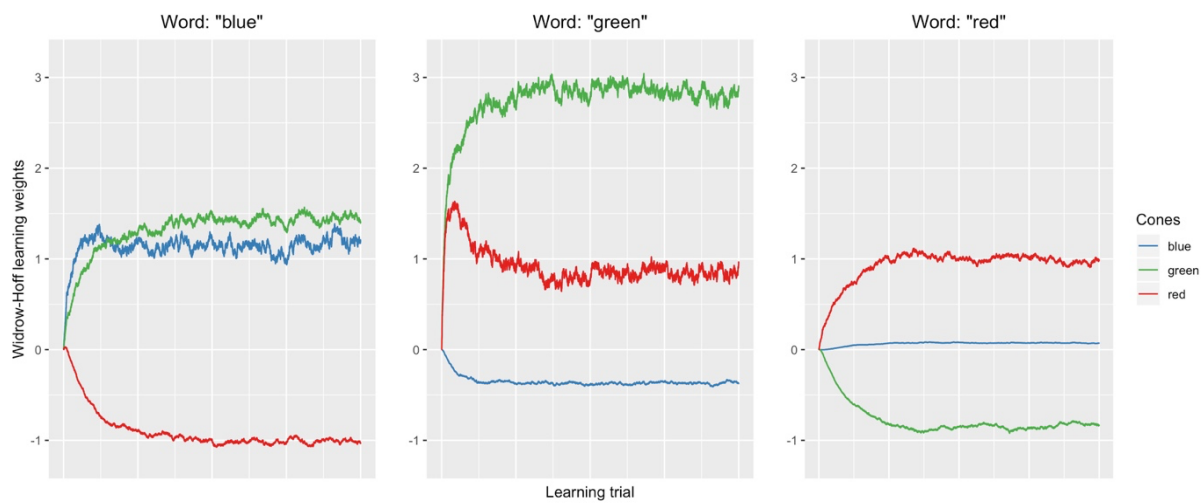

**Figure 3.** Development of learning weights from three input cues (probability values under Normal curve given randomly generated wavelengths in the range of the visible spectrum), and three outcomes (colour labels – words), in 100,000 learning trials.

Interestingly, this opens up new perspectives on a puzzling tendency in the world of colour naming. According to the seminal study by Berlin and Kay (1969), languages adopt new colour terms as they evolve, and this would happen in a strict (now criticized) chronological sequence whereby first green (Stage III or IV) and then blue (Stage V) is added to the lexicon. In many languages, the labels for blue and green remain *co-lexified*, that is, expressed using one cover term. Linguists tend to refer to the co-lexification of green and blue with the label "grue". Different hypotheses have been put forward as to why red emerges before green and green emerges before blue, or for why green and blue end up being co-lexified, but the learnability of the labels given the perceptual input have not yet been brought to bear on this question. However, as Figure 3 shows, red is the label that is straightforwardly learned from perceptual input as there is no competing perceptual activation; green experiences some competition from red, but the perceptual activation for green surpasses that for red threefold; finally, blue is always co-activated with green, and learning blue would require assuming that this label applies when green is also activated perceptually.

## 5.3. Identifying constructions from reading behaviour data

Section 5.2 serves as illustrative example of modelling simulated data to explore the complex behaviour of learning colour labels. As we pointed out, this exercise required several simplifying assumptions. Critics may object that none of these simulated results would hold in the light of the real complexity of colour labelling. Section 5.1, however, provided a summary of more than promising results for Widrow-Hoff-like learning measures in predicting human behavioural responses as collected in controlled experiments on language. In what follows we will challenge the Widrow-Hoff rule to learn and solve a 9-level classification problem, using human behavioural measures as numeric input cues and, as outcomes, 3x3 categories that combine a particular experimental manipulation from a visual-world experiment.

Details of this experiment are presented in Divjak, Milin, and Medimorec (2020). In brief, the eye movements of Native English speakers were tracked while they explored a static visual scene. The specific experimental manipulation consisted of exploring these images 'naturalistically 'or after hearing a sentence describing the scene by using a typical or atypical construction: Preposition (moveable item first [typical] or immovable item first [atypical]), Voice (active [typical] or passive [atypical]), and Dative (noun phrase [typical] or prepositional phrase [atypical]). That gave 9 possible situations in total or three per language construction (naturalistic, typical, atypical).

The Widrow-Hoff training consisted of 141 trials, corresponding to the number of actual experimental trials, with participants 'Average Pupil Size as input cues and correct category out of nine possible combinations as learning outcomes. The learning event would then be, for example:

| Trial | CUES (participants' Avg. Pupil Size) | | | | OUTCOMES (9-level exp. manipulation) | | | | |
|---|---|---|---|---|---|---|---|---|---|
| | part1 | part2 | … | part60 | prep. natural | prep. typical | … | voice atypical | … |
| **voice20r** | 1159.60 | 731.00 | … | 996.5 | 0 | 0 | 0 | 1 | 0 |

Note that the Widrow-Hoff rule, additionally, had to handle missing pupil sizes (NAs) as, due to counterbalancing, participants did not see each picture three times (for details see Divjak et al., 2020). For the purpose of training, we set all NA Pupil Size values to 0 and scaled (i.e., z-transformed) these cues to avoid multiplications of very large numbers. We randomized the order of trials and allowed Widrow-Hoff to update the learning weights from cues (Average Pupil Size) to outcomes (picture viewing situation) in one pass, with the learning rate set to $\gamma = 0.1$.

To test the performance of Widrow-Hoff weights we ran Gradient Boosting Machines (GBMs) with the **gbm** package (Greenwell, Boehmke, Cunningham, & GBM Developers, 2019) in the **R** software environment (R Core Team, 2019) to compare the performance of WH.Weights against that of the original input cues variable – AvgPupilSize of participants – as predictors for a 9-level (multinomial) classification problem. These 9 categories represented the combination of the two experimental factors (Preposition.naturalistic, Preposition.typical, Preposition.atypical, Voice.naturalistic, Voice.typical, Voice.atypical, Dative.naturalistic, Dative.typical, Dative.atypical). The GBM results (additionally 5-fold cross-validated) showed that WH.Weights was the more important predictor compared to AvgPupilSize: 67% vs. 33% of relative informativity for predicting the 9 categories. When predicting the three Constructions (Preposition, Voice, Dative) the results were again comparatively in favour of WH.Weights (55% vs. 45%). The difference came out similar for classifying TypeOfViewing (naturalistic, typical, atypical), again in favour of WH.Weights (65% vs. 35%).

The results of this case study show that Widrow-Hoff learning weights perform well as covariates (i.e., numerical, continuous predictor variables) in a multi-level classification problem. It is striking that the weights perform much better than the original (also numeric) input learning cues, as if the Widrow-Hoff rule 'distils 'the essence of the signal that is present in the input and separates it from to-be-expected noise.

This observation opens up new avenues for exploration of the emergence of rule-like behaviour in language users. Individual differences aside, language users by and large converge on a similar set of 'rules'. This happens in the face of considerable variation in language use and despite the lack of explicit error correction or language learning instruction (Pinker, 1989) during the first and arguably crucial years of language development. In fact, it has been argued that the prefrontal brain structures required for rule-based learning take until adolescence to mature (Huttenlocher, 1979; but see Werchan, Collins, Frank, & Amso, 2016). An algorithm that distils the essence from the noise would provide a plausible, yet as of yet unexplored, alternative.

## 5.4. Using Widrow-Hoff learning weights as embeddings for deep(er) learning

With the growing success of deep neural networks across domains as divergent as computer vision and machine translation (cf., Krizhevsky, Sutskever, & Hinton, 2012), those working on language have been tempted to adopt these methods for their purposes. The biggest obstacle, however, has been in representing textual information in a manner that can be fed into deep neural networks, which require their input to be numeric (Jurafsky & Martin, 2000). While images and sounds are trivially easy to translate into numbers, this is not the case for text. This challenge has been met in different ways, leading to a variety of numeric (vector) representations of words, including **word2vec** (Mikolov, Sutskever, Chen, Corrado, & Dean, 2013), **GloVe** (Pennington, Socher, & Manning, 2014), and more recently representations that depend on the context a word occurs in, such as **BERT** (Devlin, Chang, Lee, & Toutanova, 2018).

In this case study, we compare Widrow-Hoff learning weights with GloVe, a comparable, non-contextual embedding, on the downstream task of verb prediction in Russian. This task involves the prediction of a verb's 'meaning 'labels in a given sentence; for the purpose of this study, well-researched semantic groups of near-synonymous verbs expressing attempt and perception were selected (Divjak & Gries, 2006; Divjak, 2015). The verb labels associated with each lemma are listed in Table 2. Note that we have opted for using labels instead of lemmata so as to simplify the problem from a 12 way classification problem to a 4 way classification problem, thus speeding up training while retaining the overall characteristics of the prediction task.

**Table 2.** List of targeted Russian verbs and their descriptive label.

| Verb | Label |
| --- | --- |
| попробовать | TRY |
| попытаться | TRY |
| постараться | TRY |
| пробовать | TRY |
| пытаться | TRY |
| стараться | TRY |
| послушать | LISTEN |
| слушать | LISTEN |
| посмотреть | LOOK |
| смотреть | LOOK |
| трогать | TOUCH |
| тронуть | TOUCH |

For this case study we devised two types of the Widrow-Hoff embeddings. We made use of the Araneum Russicum (Benko, 2014; Benko & Zakharov, 2016), a large web-based corpus comprising of 1.2 billion word tokens in more than 71 million sentences. First, the corpus was filtered to retain only sentences with permissible Russian words, consisting only of Russian characters. Numerals and punctuation were removed, and all letters were lowercased. Next, we pre-processed the sentences further to obtain the learning events for Widrow-Hoff training. Actual training was done with the **pyndl** library (Sering, Weitz, Künstle, & Schneider, 2017) in the **Python** programming environment. The library implements the Rescorla-Wagner learning rule, a categorical variant of the more general Widrow-Hoff rule.

Learning events were built by moving a window of four words by one word at each step; the window captured the target word, two preceding words, and one following word. For example, for a sentence such as "Jack was a fearless but funny pirate.", the first learning event would not have any preceding context. "Jack" would be the target word followed by "was". Similarly, the fourth learning event would be "was, a" (preceding words), "fearless" (the target), "but" (the following word). In other words, the initial word as target would have no preceding context and the final word would have no following context. The total number of learning events was 941 million. The learning rate was set to $\gamma = 0.001$. The resulting matrix of learning weights gave us a dictionary of 40,000 Russian words, represented as 40,000-long numeric vectors.

Finally, from the resulting matrix, we built two types of Widrow-Hoff based embeddings by constraining the words 'vector representation to a length of 300. To achieve this, we rely on the notion of Diversity from the NDL framework (for details, see Milin, Feldman, et al., 2017). Practically, for each word in the dictionary, we calculated its absolute length (1-norm; Diversity in NDL terminology) and then selected 300 context words such that they are either (1) context words with the highest Diversity as vector-point representations (i.e., context words with the highest absolute length); (2) context words randomly selected across the range of Diversity values (i.e., a random selection from the range of absolute lengths). This gave us two types of Widrow-Hoff embeddings which we evaluate using a standard evaluation method followed by the task of predicting Russian verbs.

The most common method of evaluating the effectiveness of word embeddings uses Spearman's rank correlation coefficient between the similarity score assigned by human annotators and that generated by the embeddings. We use the human annotated dataset SimLex-999 (Hill, Reichart, & Korhonen, 2015), which is a gold standard resource. Table 3 lists the scores achieved on the two Widrow-Hoff embeddings and GloVe embeddings. As both GloVe and Widrow-Hoff embeddings lack some of the words present in the evaluation dataset, we perform this experiment twice – once with missing words ignored (listed in Table 3 as OOV) thus limiting ourselves only to those words within the embeddings (listed in Table 3 as INV) and a second time by limiting ourselves to the words in the evaluation data that are present across all three embeddings. It should be noted that the Widrow-Hoff Sample Diverse embeddings perform significantly better than GloVe. However, our experiments also show that the Sample Diverse embeddings are very sensitive to the (random) selection of context words and vary considerably based on this selection. It should also be noted that the Widrow-Hoff embeddings have a larger number of out of vocabulary words.

**Table 3.** The similarities (rank-correlations) between human annotators 'scores and scores based on the three types of embedding.

| Embedding | OOV | INV | Score |
|-----------|-----|-----|-------|
| GloVe | 226 | 739 | 0.1257 |
| WHMD | 372 | 593 | 0.0731 |
| WHSD | 372 | 593 | 0.2413 |
| | | | |
| GloVe | 0 | 551 | 0.1842 |
| WHMD | 0 | 551 | 0.0899 |
| WHSD | 0 | 551 | 0.2463 |

The model we utilised for our case study - the task of predicting 4 types of Russian verbs - made use of a three-layer hierarchical LSTM, which feeds into a conditional random field (CRF) to capture dependencies (cite Liu et al., 2018). This is the standard model used in Natural Language Processing to label individual words in a sentence. It is especially effective due to the ability of LSTMs to capture long dependencies. We trained the model for 20 epochs with (a) The Russian version of the the pre-trained GloVe embeddings generated by reconstructing the English version by use of the TED talks parallel text corpora (https://www.aclweb.org/anthology/P16-1190.pdf), (b) Widrow-Hoff Most Diverse embeddings (WHMD), (c) Widrow-Hoff Sample Diverse embeddings (WHSD), and (d) randomly initialised embeddings that we used as baseline. In the case of the randomly initialised word embeddings, we allowed our model to fine-tune the embeddings, whereas, in the case of the other three types of word embeddings, we did not allow such fine-tuning, so as to better capture the effectiveness of the embeddings themselves.

As neural models tend to be extremely sensitive to initialisation, which is random, we completed five runs for each of the four embeddings. Additionally, as is standard practice, we optimised the model on a development set and evaluated the embeddings 'performance on a different test dataset, to avoid over-optimising the training process. We provide additional details of the parameters used during training in Table 4 and present our results in Table 5.

**Table 4.** LSTM CRF Model Hyperparameters.

| | |
|---|---|
| **Batch Size:** | 50 |
| **Ignore Case:** | TRUE |
| **Clip grad:** | 5 |
| **Drop out:** | 0 |
| **Hidden:** | 100 |
| **LSTM layers:** | 3 |
| **Learning rate (LR):** | 0.015 |
| **LR Decay:** | 0.05 |
| **Momentum:** | 0.9 |
| **Patience:** | 15 |

**Table 5.** Effectiveness results for the four types of LSTM embeddings.

| Embeddings | Development set | | | | Test set | | | |
|---|---|---|---|---|---|---|---|---|
| | Max F1 | Min F1 | Std. Dev. | Avg. | Max F1 | Min F1 | Std. Dev. | Avg. |
| **Random** | 0.500 | 0.385 | 0.044 | 0.430 | 0.453 | 0.376 | 0.032 | 0.406 |
| **GloVe** | 0.790 | 0.753 | 0.017 | 0.761 | 0.716 | 0.605 | 0.042 | 0.672 |
| **WHMD** | 0.667 | 0.580 | 0.034 | 0.630 | 0.704 | 0.543 | 0.062 | 0.598 |
| **WHSD** | 0.728 | 0.568 | 0.065 | 0.644 | 0.605 | 0.494 | 0.043 | 0.562 |

The results indicate that the two Widrow-Hoff based embeddings (WHMD, WHSD) perform with an improvement of between 15% and 20% in comparison with the random baseline. However, the GloVe embeddings do perform better than both Widrow-Hoff embeddings. Whilst WHSD performs better on word similarity, GloVe performs better on next word prediction which has less to do with similarity and more to do with word analogy. WHSD embeddings also suffer from a significantly smaller vocabulary. We thus conclude that WHSD embeddings, with a larger coverage, will be a better choice than GloVe for tasks that depend on word similarity.

We note, however, that the average F1 score on the development set is very close to that of the test set for both Widrow-Hoff embeddings, but unlike that of GloVe where the difference is much higher. We also note that the highest score achieved by the Most Diverse embeddings is very close to that of GloVe although the Most Diverse embeddings do seem to be more sensitive to initialisation and the average is lower with a correspondingly high standard deviation.

## 6. "Much Ado About" the amount of data, (non) linearity, convergence, overfitting

We have already mentioned that the main point of criticism of the early artificial neural models, in general, and Widrow-Hoff's (1960) ADELINE, in particular, was their applicability to linearly separable problems only (cf., Minsky & Papert, 1988). Widrow and Lehr's (1990) countered that limits on ADELINE's *processing capacity* unleash its *capacity to generalize* and, thus, learn. Importantly, however, even the tiniest network of two ADELINES (i.e., a MADELINE) would be able to handle non-linearities in the input and constitute what are today called Fully Connected Neural Networks. Elemental units (ADELINEs) could be set in parallel (e.g., Widrow & Lehr, 1990, p. 1420) or serially (e.g., Roy & Chakraborty, 2013, p.243) to solve exclusive-or (XOR) and similar problems.

Milin, Feldman, et al. (2017) analysed in greater details the processing capacity of the Rescorla-Wagner rule (identical to the Widrow-Hoff rule, according to Rescorla, 2008). They showed that the rule's performance against a simulated problem that is set as a non-linear classification problem is, in fact, much more similar to the performance of Linear Regression with Lasso regularization and, overall, much better than the performance of Simple Linear Regression. The authors argued that the performance of a model cannot be determined only by the algorithm but must also take into account the representation of the input-output space. The logic is that underlying Support Vector Machines that re-project the data into a space with more dimension that allows for linear separation, or when deep(er) learning networks re-parametrize the input with hidden units which, in a final step, also approach to a linear solution (see Baayen & Hendrix, 2017 for detailed discussion).

Another important point is that of the data itself. As Widrow and Hoff (1960) argued, the performance of the model depends crucially on the amount of experience – the data. What they did not specify but nevertheless implicitly assumed is that the experience or data ought to be *new*. If we recycle the same data in repeated training runs, we are forcing the model to 'believe 'that data too much: the more it commits to particular data, the less it will be able to deal with *newsiness* and hence demonstrate learning and its ability to generalize.

In many practical applications it is often customary to allow multiple runs over the same data to encourage convergence. A balance must be found, however, between convergence and overfit. To understand this problem better we trained Widrow-Hoff on a simulated dataset, using a multivariate Gaussian number generator, with four numeric variables: a criterion and three predictors (Means: $y = 0.0, x_1 = 5.0, x_2 = 10.0, x_3 = -2.0$; Standard deviations: $y = 1.75$, $x_1 = 1.5$, $x_2 = 1.25$, $x_3 = 1.0$) with the following inter-correlations (set in such a way to have high correlations between the criterion and the three predictors, while keeping the correlations between predictors low to control for multicollinearity):

|       | y    | $x_1$ | $x_2$ | $x_3$ |
|-------|------|------|------|------|
| y     | -    |      |      |      |
| $x_1$ | 0.66 | -    |      |      |
| $x_2$ | 0.62 | 0.15 | -    |      |
| $x_3$ | 0.58 | 0.10 | 0.12 | -    |

For the reference model we ran Multiple Linear Regression ($R^2_{Adjusted} = 0.93$; $F = 4504$; $df = 3 and 996$; $p-value < 0.00001$). Then, we ran three different Widrow-Hoff learning sessions on the same data, adding a bias input of $1.0$ (bias input in the terminology of Widrow & Hoff, 1960 and Widrow & Lehr, 1990 corresponds to the Intercept in Linear Regression) and using a very small learning rate of $\gamma = 0.0001$: (1) single run learning; (2) 10,000 repeated learning runs; (3) 10,000 repeated runs per learning trial, which were sorted in ascending order of the outcome values prior to learning. Table 6 summarizes the results.

The results show very similar values for Linear Regression and Widrow-Hoff learning with 10,000 repetition runs. Thus, with many repetitions and a conveniently small learning rate that allows for gradual change (i.e., minimal disturbance), the Widrow-Hoff rule indeed achieves LMS (for this particular simulated and randomized dataset). However, the same number of expositions but with 10,000 repetitions of each trial with trials sorted in ascending order of outcome values, gives very different results. Further increases in the number of repeated runs (e.g., to 100,000 repetitions) does not affect the weight values for these two learning sessions and we can conclude that they have indeed converged. Conversely, with a very small, miniscule learning rate, a single learning run actually shows that little has been learned but, arguably, not much less than with sorted repeated trials.

Crucially, this example confirms that the amount of new experience and how that experience is ordered matters and factors into the rate of informative and non-informative examples (cf., Rescorla & Wagner, 1972; Rescorla, 1988; Ramscar et al., 2010; Milin et al., under review). In the end, when the ordering effect is annulled by randomization and with a sufficient number of repetitions, the Widrow-Hoff rule will converge to the maximum likelihood estimate of the learning weight ($W_i$; see Gershman, 2015), at the risk of overfitting for a particular dataset.

**Table 6.** Estimates and weights from Multiple Linear Regression and three Widrow-Hoff learning session using three numeric predictors (cues) and single numeric criterion (outcome).

| | Estimates / Weights | | | |
| --- | --- | --- | --- | --- |
| | Linear Regression | Single WH run | 10K WH runs | 10K sorted WH runs |
| Intercept / Bias | -8.1325 | -0.0161 | -8.1232 | -0.0141 |
| $x_1$ | 0.6270 | 0.0856 | 0.6242 | 0.2089 |
| $x_2$ | 0.6632 | -0.0292 | 0.6575 | 0.2954 |
| $x_3$ | 0.8094 | 0.1250 | 0.8108 | 0.0622 |

Evert and Arppe (2015) took an applied statistical point of view and discussed the performance of Rescorla-Wagner rule in terms of *success* or *failure* to converge to the estimates of linear regression. We, however, believe that such an 'ideal 'is, in fact, faux and inequitable to the genuine strengths of simple incremental error-correction learning models (with Rescorla-Wagner, as we pointed out, being evaluated as identical to the Widrow-Hoff rule by Rescorla, 2008 himself): they are developed for the purpose of illustrating the process of learning and adaptive behaviour (cf., Widrow & Hoff, 1960, p. 97), with a *by-design* limited capacity in performance traded for an enhanced capacity to generalize (cf., Widrow & Lehr, 1990, p. 1422). Nevertheless, it has been shown that these models will converge to the maximum likelihood estimate if they are fed with large amounts of data in a randomized order. However, healthy living beings do not experience an unstructured (or non-autocorrelated) and endless stream of data. Arguably, life experience becomes by definition gradually more 'meaningful 'and structured (cf., James, 1890), allowing for the evolution of a particular type of "learning machines" (Poggio, 2012, p. 1919)?

**7. Concluding remarks**

In this paper we introduced a computational implementation of one of the earliest neurally inspired models of learning, the Widrow-Hoff (1960) rule. Deciding to apply this particular rule implies agreeing with several assumptions: data is made available to the learner in a piece-meal fashion and learning 'evolves 'over time by minimizing the number of erroneous predictions (viz. error-correction learning); *achieved* performance is *optimal* performance (given the constrains set by the assumptions) which is not (and, possibly, never is) the same as *error-free* performance.

These assumptions fit the development of language knowledge rather well. After all, during our linguistic development we are not bombarded with infinite amounts of data presented to us in a random order. Also, often, there is no one correct outcome. Yet, modelling learning over large datasets with many inputs and outputs, as is the case for language data, poses implementational challenges. These can be overcome by making full use of parallelisation options, the computing power offered by GP-GPUs, and software optimized for matrix and vector operations.

Our case studies showed that we achieve surprising results. Widrow-Hoff does an exceptional job modelling the learnability or predictability of labels given the perceptual (colours) or textual (near synonyms) input. In fact, in the latter domain, the performance of the simple learning rule is on a par with that of state-of-the art data-preprocessing techniques which constitute the input for models of deep learning. Crucially, however, the present results reflect (or constrain) learning that is biologically or cognitively plausible. We have also shown that Widrow-Hoff learning weights help filter the signal from noisy input. In a sense, they detect strong regularities in highly variable input. This property of the Widrow-Hoff rule makes it eminently suited for the investigation of the emergence of abstractions in language.

# References

Arnon, I., & Ramscar, M. (2012). Granularity and the acquisition of grammatical gender: How order-of-acquisition affects what gets learned. *Cognition, 122*(3), 292-305.

Baayen, R. H., & Hendrix, P. (2017). Two-layer networks, non-linear separation, and human learning. In M. Wieling, M. Kroon, G. van Noord, O. Strik, & G. Bouma (Eds.), *From Semantics to Dialectometry. Festschrift in honor of John Nerbonne* (Vol. 32, pp. 13-22). London: College Publications.

Baayen, R. H., Milin, P., Đurđević, D. F., Hendrix, P., & Marelli, M. (2011). An amorphous model for morphological processing in visual comprehension based on naive discriminative learning. *Psychological review, 118*(3), 438-481.

Benko, V. (2014). *Aranea: Yet another family of (comparable) web corpora.* Paper presented at the International Conference on Text, Speech, and Dialogue.

Benko, V., & Zakharov, V. P. (2016). Very large Russian corpora: new opportunities and new challenges. In *Computational linguistics and intellectual technologies* (pp. 79-93): Российский государственный гуманитарный университет.

Berlin, B., & Kay, P. (1969). *Basic color terms: Their universality and evolution*. Berkeley, CA: University of California Press.

Blough, D. S. (1975). Steady state data and a quantitative model of operant generalization and discrimination. *Journal of Experimental Psychology: Animal Behavior Processes, 1*(1), 3-21.

Chen, Z., Haykin, S., Eggermont, J. J., & Becker, S. (2008). *Correlative learning: a basis for brain and adaptive systems*. New Jersey: John Wiley & Sons.

Cox, D. R. (1966). Some procedures connected with the logistic qualitative response curve. In F. N. David (Ed.), *Research Papers in Statistics* (pp. 55-71). London: Wiley.

Cramer, J. S. (2002). *The Origins of Logistic Regression*. Retrieved from

Devlin, J., Chang, M.-W., Lee, K., & Toutanova, K. (2018). Bert: Pre-training of deep bidirectional transformers for language understanding. *arXiv preprint arXiv:1810.04805*.

Divjak, D. (2015). Exploring the grammar of perception: A case study using data from Russian. *Functions of Language, 22*(1), 44-68.

Divjak, D., & Gries, S. T. (2006). Ways of trying in Russian: Clustering behavioral profiles. *Journal of Corpus Linguistics and Linguistic Theory, 2*(1), 23-60.

Divjak, D., Milin, P., & Medimorec, S. (2020). Construal in language: A visual-world approach to the effects of linguistic alternations on event perception and conception. *Cognitive Linguistics, 30*(1), online first.

Dowling, J. E. (1987). *The retina: an approachable part of the brain*. Cambridge, MA: Harvard University Press.

Enquist, M., & Ghirlanda, S. (2005). *Neural networks and animal behavior* (Vol. 33). Princeton: Princeton University Press.

Enquist, M., Lind, J., & Ghirlanda, S. (2016). The power of associative learning and the ontogeny of optimal behaviour. *Royal Society open science, 3*(11), 160734.

Evert, S., & Arppe, A. (2015). *Some theoretical and experimental observerations on naive discriminative learnin.* Paper presented at the The 6th Conference on Quantitative Investigations in Theoretical Linguistics, Tübingen, Germany.

Gershman, S. J. (2015). A unifying probabilistic view of associative learning. *PLoS computational biology, 11*(11), e1004567.

Goodfellow, I., Bengio, Y., & Courville, A. (2016). *Deep Learning*. Cambridge, MA: MIT Press.

Greenwell, B., Boehmke, B., Cunningham, J., & GBM Developers. (2019). gbm: Generalized boosted regression models (Version 2.1.5). Retrieved from https://CRAN.R-project.org/package=gbm

Gureckis, T. M., & Love, B. C. (2015). Computational reinforcement learning. In J. R. Busemeyer, Z. Wang, J. T. Townsend, & A. Eidels (Eds.), *The Oxford handbook of computational mathematical psychology* (pp. 99-117). New York, NY: Oxford University Press.

Haykin, S. S. (1996). *Adaptive filter theory*. Upper Saddle River, NJ: Prentice Hall.

Haykin, S. S. (1999). Neural networks: A comprehensive foundation. In (2nd ed. ed.). London: Prentice Hall.

Hebb, D. O. (1949). *The organization of behavior*. New York: Wiley.

Hill, F., Reichart, R., & Korhonen, A. (2015). Simlex-999: Evaluating semantic models with (genuine) similarity estimation. *Computational Linguistics, 41*(4), 665-695.

Huttenlocher, P. R. (1979). Synaptic density in human frontal cortex-developmental changes and effects of aging. *Brain Res, 163*(2), 195-205.

Ivens, S. H., & Koslin, B. L. (1991). Demands for reading literacy require new accountability methods. from Touchstone Applied Science Associates

James, W. (1890). *The principles of psychology (Vol. 1)* (Vol. 474). New York: Henry Holt and Co.

Jurafsky, D., & Martin, J. H. (2000). *Speech and Language Processing: An Introduction to Natural Language Processing, Computational Linguistics, and Speech Recognition*. Upper Saddle River, NJ: Prentice Hall.

Kalman, R. E. (1960). A new approach to linear filtering and prediction problems. *Journal of basic Engineering, 82*(1), 35-45.

Kamin, L. J. (1969). Predictability, surprise, attention, and conditioning. In B. Campbell & R. Church (Eds.), *Punishment and aversive behaviour* (pp. 279-296). New York: Appleton-Century-Crofts.

Kohonen, T. (1972). Correlation matrix memories. *IEEE transactions on computers, 100*(4), 353-359.

Krizhevsky, A., Sutskever, I., & Hinton, G. E. (2012). *Imagenet classification with deep convolutional neural networks.* Paper presented at the Advances in neural information processing systems.

Kruschke, J. K. (2008). Bayesian approaches to associative learning: From passive to active learning. *Learning & behavior, 36*(3), 210-226.

Liu, L., Shang, J., Ren, X., Xu, F. F., Gui, H., Peng, J., & Han, J. (2018). *Empower sequence labeling with task-aware neural language model.* Paper presented at the Thirty-Second AAAI Conference on Artificial Intelligence.

Lucky, R. W. (1965). Automatic equalization for digital communication. *Bell System Technical Journal, 44*(4), 547-588.

McClelland, J. L., & Rumelhart, D. E. (1986). *Parallel distributed processing: Explorations in the microstructure of cognition* (Vol. 1). Cambridge: MIT press.

McCulloch, W. S., & Pitts, W. (1943). A logical calculus of the ideas immanent in nervous activity. *The bulletin of mathematical biophysics, 5*(4), 115-133.

Mikolov, T., Sutskever, I., Chen, K., Corrado, G. S., & Dean, J. (2013). *Distributed representations of words and phrases and their compositionality.* Paper presented at the Advances in neural information processing systems.

Milin, P., Divjak, D., & Baayen, R. H. (2017). A Learning Perspective on Individual Differences in Skilled Reading: Exploring and Exploiting Orthographic and Semantic Discrimination Cues. *Journal of*

*Experimental Psychology: Learning, Memory, and Cognition, 43*(11), 1730-1751. doi:10.1037/xlm0000410

Milin, P., Feldman, L. B., Ramscar, M., Hendrix, P., & Baayen, R. H. (2017). Discrimination in lexical decision. *PloS one, 12*(2), e0171935.

Milin, P., Nenadić, F., & Ramscar, M. (under review). Approaching text genre: How contextualized experience shapes task-specific performance. *Scientific Study of Literature*.

Minsky, M. L., & Papert, S. (1988). *Perceptrons : an introduction to computational geometry*.

Niv, Y. (2009). Reinforcement learning in the brain. *Journal of Mathematical Psychology, 53*(3), 139-154.

Pennington, J., Socher, R., & Manning, C. D. (2014). *Glove: Global vectors for word representation.* Paper presented at the Proceedings of the 2014 conference on empirical methods in natural language processing (EMNLP).

Pinker, S. (1989). *Learnability and cognition: The acquisition of argument structure*. Cambridge, MA: MIT Press.

Poggio, T. (2012). The levels of understanding framework, revised. *Perception, 41*(9), 1017-1023.

R Core Team. (2019). R: A language and environment for statistical computing (Version 3.6.2). Vienna, Austria: R Foundation for Statistical Computing. Retrieved from https://www.R-project.org/

Ramscar, M., Yarlett, D., Dye, M., Denny, K., & Thorpe, K. (2010). The Effects of Feature- Label- Order and Their Implications for Symbolic Learning. *Cognitive Science, 34*(6), 909-957. doi:10.1111/j.1551-6709.2009.01092.x

Rescorla, R. A. (1968). Probability of shock in the presence and absence of CS in fear conditioning. *Journal of comparative and physiological psychology, 66*(1), 1-5.

Rescorla, R. A. (1988). Pavlovian conditioning: It's not what you think it is. *American psychologist, 43*(3), 151-160.

Rescorla, R. A. (2003). Contemporary study of Pavlovian conditioning. *The Spanish journal of psychology, 6*(2), 185-195.

Rescorla, R. A. (2008). Rescorla-Wagner model. *Scholarpedia, 3*(3), 2237.

Rescorla, R. A., & Wagner, R. A. (1972). A theory of Pavlovian conditioning: variations in the effectiveness of reinforcement and non-reinforcement. In H. Black & W. F. Proksay (Eds.), *Classical conditioning II* (pp. 64-99). New York, NY: Appleton-Century-Crofts.

Rosenblatt, F. (1958). The perceptron: a probabilistic model for information storage and organization in the brain. *Psychological review, 65*(6), 386-408.

Roy, S., & Chakraborty, U. (2013). *Introduction to Soft computing: Neuro-Fuzzy and Genetic Algorithms*: Pearson Education India.

Schubert, E. F. (2006). *Light-Emitting Diodes*: E. Fred Schubert.

Sering, K., Weitz, M., Künstle, D.-E., & Schneider, L. (2017). Pyndl: Naive discriminative learning in python. Retrieved from https://doi.org/10.5281/zenodo.597964

Sondhi, M. M. (1967). An adaptive echo canceller. *Bell System Technical Journal, 46*(3), 497-511.

Sutton, R. S. (1992). *Gain adaptation beats least squares.* Paper presented at the Proceedings of the 7th Yale workshop on adaptive and learning systems.

Sutton, R. S., & Barto, A. G. (1990). Time-derivative models of pavlovian reinforcement. In M. Gabriel & J. Moore (Eds.), *Learning and computational neuroscience: Foundations of adaptive networks* (pp. 497-537). Cambridge, MA: The MIT Press.

Trimmer, P. C., McNamara, J. M., Houston, A. I., & Marshall, J. A. (2012). Does natural selection favour the Rescorla–Wagner rule? *Journal of theoretical biology, 302*, 39-52.

Werchan, D. M., Collins, A. G., Frank, M. J., & Amso, D. (2016). Role of prefrontal cortex in learning and generalizing hierarchical rules in 8-month-old infants. *Journal of Neuroscience, 36*(40), 10314-10322.

Widrow, B. (1959). *Adaptive sampled-data systems—a statistical theory of adaptation.* Paper presented at the IRE Wescon Convention Record.

Widrow, B. (1960). *Adaptive sampled-data systems.* Paper presented at the Proceedings of the First International Congress of the International Federation of Automatic Control.

Widrow, B. (1962). Generalization and Information Storage in Networks of ADALINE Neurons. In M. C. Yovitz, G. T. Jacobi, & G. D. Goldstein (Eds.), *Self Organizing Systems* (pp. 435-461). Washington: Spartan Books.

Widrow, B. (1987). *The original adaptive neural net broom-balancer.* Paper presented at the IEEE International Symposium on Circuits and Systems.

Widrow, B., & Hoff, M. E. (1960). *Adaptive switching circuits.* Paper presented at the WESCON Convention Record Part IV.

Widrow, B., & Lehr, M. A. (1990). 30 years of adaptive neural networks: perceptron, madaline, and backpropagation. *Proceedings of the IEEE, 78*(9), 1415-1442.

Widrow, B., Mantey, P., Griffiths, L., & Goode, B. (1967). Adaptive antenna systems. *Proceedings of the IEEE, 55*(12), 2143-2159.

Widrow, B., & Stearns, S. D. (1985). *Adaptive signal processing*. Upper Saddle River, NJ: Prentice-Hall.